\pdfoutput=1

\documentclass[11pt]{article}

\usepackage{EMNLP2022}

\usepackage{times}
\usepackage{latexsym}

\usepackage[T1]{fontenc}

\usepackage[utf8]{inputenc}

\usepackage{microtype}

\usepackage{inconsolata}

\usepackage{amsmath}
\usepackage{amssymb}
\usepackage{pifont}
\usepackage{graphicx}
\usepackage{graphics}
\usepackage{hyperref}
\usepackage{url}
\usepackage{multirow}
\usepackage{subfigure}
\usepackage{colortbl}
\usepackage{makecell}
\usepackage{booktabs}
\usepackage{arydshln}
\usepackage{enumitem}

\definecolor{mediumelectricblue}{rgb}{0.01, 0.31, 0.59}
\usepackage{booktabs}

\newcommand{\electricblue}[1]{\textcolor{mediumelectricblue}{#1}}

\newcommand{\method}{\textsc{PALT}}
\newcommand{\methodLarge}{\method$_{\rm LARGE}$}

\newcommand{\prompt}{knowledge prompt encoder}
\newcommand{\calibration}{knowledge calibration encoder}
\newcommand{\pl}{parameter-lite}
\newcommand{\correctmark}{\textcolor{green}{\ding{51}}}
\newcommand{\wrongmark}{\textcolor{red}{\ding{55}}}
\newcommand{\texthrt}[1]{\textsl{#1}}
\newcommand{\textspt}[1]{\texttt{#1}}
\newcommand{\comm}[1]{}

\newif\ifshowcomment
\showcommentfalse

\ifshowcomment
\newcommand{\jianhao}[1]{}
\newcommand{\todo}[1]{\textcolor{red}{[TODO: #1]}}
\newcommand{\yuanye}[1]{}
\newcommand{\focus}[1]{}
\else
\newcommand{\jianhao}[1]{}
\newcommand{\yuanye}[1]{}
\newcommand{\todo}[1]{}
\newcommand{\focus}[1]{}
\fi

\title{\method: Parameter-Lite Transfer of Language Models for \\ Knowledge Graph Completion}

\author{Jianhao Shen\textsuperscript{1}, Chenguang Wang\textsuperscript{2}$^{\dagger}$, Ye Yuan\textsuperscript{1}, Jiawei Han\textsuperscript{3}\\
{\bf Heng Ji\textsuperscript{3}}, {\bf Koushik Sen\textsuperscript{4}}, {\bf Ming Zhang\textsuperscript{1}}$^{\dagger}$, {\bf Dawn Song\textsuperscript{4}}$^{\dagger}$\\
\textsuperscript{1}Peking University, \textsuperscript{2}Washington University in St. Louis, \\
\textsuperscript{3}University of Illinois at Urbana-Champaign, \textsuperscript{4}UC Berkeley\\
\texttt{\{jhshen,yuanye\_pku,mzhang\_cs\}@pku.edu.cn}, \texttt{chenguangwang@wustl.edu},\\ \texttt{\{hanj,hengji\}@illinois.edu}, \texttt{\{ksen,dawnsong\}@berkeley.edu}
}

\begin{document}
\maketitle
 
\def\thefootnote{$^\dagger$}\footnotetext{Corresponding authors.}
\def\thefootnote{\arabic{footnote}}
 
\begin{abstract}
\todo{fix figure, fix consistency (short (e.g., lm vs language model, kg vs knowledge graph, fact instead of triplet, Transformer instead of transformer) including appendix, add appendix reference), improve appendix writing, check numbers, add meta review papers, fix grammars and other issues (e.g., extra space), remove all vspace throughout the paper, send overleaf to all authors}
This paper presents a parameter-lite transfer learning approach of pretrained language models (LM) for knowledge graph (KG) completion. Instead of finetuning, which modifies all LM parameters, we only tune a few new parameters while keeping the original LM parameters fixed. We establish this via reformulating KG completion as a ``fill-in-the-blank'' task, and introducing a parameter-lite encoder on top of the original LMs. We show that, by tuning far fewer parameters than finetuning, LMs transfer non-trivially to most tasks and reach competitiveness with prior state-of-the-art approaches. For instance, we outperform the fully finetuning approaches on a KG completion benchmark by tuning only 1\% of the parameters.\footnote{\label{ft:opensource}The code and datasets are available at \url{https://github.com/yuanyehome/PALT}.}
\end{abstract}

\section{Introduction}
Pretrained language models (LM) such as BERT and GPT-3 have enabled downstream transfer~\cite{BERT, brown_language_2020}. Recent studies~\cite{LAMA, lm_know, he2021empirical} show that the implicit knowledge learned during pretraining is the key to success. Among different transfer learning techniques~\cite{shin-etal-2020-autoprompt, liu_pre-train_2021, liu_gpt_2021, pmlr-v97-houlsby19a, BERT}, finetuning is the de facto paradigm to adapt the knowledge to downstream NLP tasks. Knowledge graph (KG) completion is a typical knowledge-intensive application. For example, given a fact \texthrt{(Chaplin, profession, \_\_)} missing an entity, it aims to predict the correct entity ``screenwriter''. This task provides a natural testbed to evaluate the knowledge transfer ability of different transfer learning approaches. 

Finetuning~\cite{KGBERT, lass} has been recently adopted to advance the KG completion performance. However, it presents two fundamental limitations. First, finetuning is computationally inefficient, requiring updating all parameters of the pretrained LMs. This ends up with an entirely new model for each KG completion task. For example, storing a full copy of pretrained BERT$_{\rm LARGE}$ (340M parameters) for each task is non-trivial, not to mention the billion parameter LMs. Second, the finetuning approaches often rely on task-specific architectures for various KG completion tasks. For instance, KG-BERT~\cite{KGBERT} designs different model architectures to adapt a pretrained BERT to different tasks. This restricts its usability in more downstream tasks.

In this work, we enable parameter-lite transfer of the pretrained LMs to knowledge-intensive tasks, with a focus on KG completion. As an alternative to finetuning, our method, namely \method, tunes no existing LM parameters. We establish this by casting the KG completion into a ``fill-in-the-blank'' task. This formulation enables eliciting general knowledge about KG completion from pretrained LMs. By introducing a parameter-lite encoder consisting of a few trainable parameters, we efficiently adapt the general model knowledge to downstream tasks. The parameters of the original LM network remain fixed during the adaptation process for different KG completion tasks. In contrast to finetuning which modifies all LM parameters, \method\ is lightweight. Instead of designing task-specific model architectures, \method\ stays with the same model architecture for all KG completion tasks that we evaluate. 

The contributions are as follows:
\begin{itemize}
    \item We propose parameter-lite transfer learning for pretrained LMs to adapt their knowledge to KG completion. The reach of the results is vital for broad knowledge-intensive NLP applications.
    \item We reformulate KG completion as a ``fill-in-the-blank'' task. This new formulation helps trigger pretrained LMs to produce general knowledge about the downstream tasks. The new formulation implies that the KG completion can serve as a valuable knowledge benchmark for pretrained LMs, in addition to benchmarks such as LAMA~\cite{LAMA} and KILT~\cite{kilt}.
    \item We introduce a parameter-lite encoder to specify general model knowledge to different KG completion tasks. This encoder contains a few parameters for providing additional context and calibrating biased knowledge according to the task. The module is applicable to other deep LMs.
    \item We obtain state-of-the-art or competitive performance on five KG completion datasets spanning two tasks: link prediction and triplet classification. We achieve this via learning only 1\% of the parameters compared to the fully finetuning approaches. In addition, compared to task-specific KG completion models, \method\ reaches competitiveness with a unified architecture for all tasks.
\end{itemize}
\section{\method}
We propose \pl\ transfer learning, called \method, as an alternative to finetuning for knowledge graph (KG) completion. Instead of finetuning which modifies all the language model (LM) parameters and stores a new copy for each task, this method is lightweight for KG completion, which keeps original LM parameters frozen, but only tunes a small number of newly added parameters. The intuition is that LMs have stored factual knowledge during the pretraining, and we need to properly elicit the relevant knowledge for downstream tasks without much modification to the original LMs. To do so, \method\ first casts KG completion into a ``fill-in-the-blank'' task (Sec.~\ref{sec:zs}), 
and then introduces a \pl\ encoder consisting of a few trainable parameters, while parameters of the original network remain fixed (Sec.~\ref{sec:pl}). The overall architecture of \method\ is shown in Figure~\ref{fig:model}. 

\subsection{Knowledge Graph Completion as Fill-in-the-Blank}
\label{sec:zs}

\begin{figure}[]
    \centering
    \includegraphics[width=\linewidth]{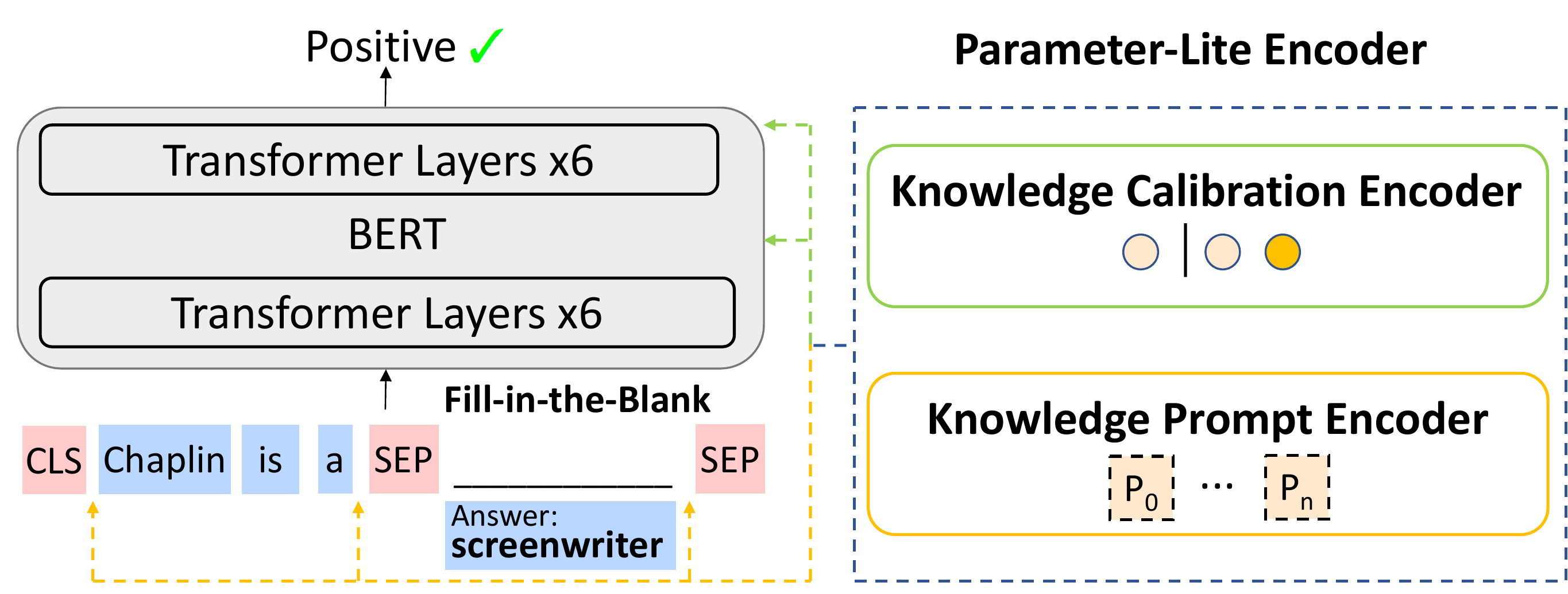}
    \caption{Summary of our approach \method. Compared to finetuning, \method\ is a parameter-lite alternative to transfer the knowledge that pretrained language models know about knowledge graph completion. Our approach first casts knowledge graph completion into a fill-in-the-blank task. This formulation enables pretrained language models to produce general knowledge for knowledge graph completion. By introducing a few trainable parameters via a parameter-lite encoder (in 
   the dashed box), \method\ further adapts the general knowledge in language models to different knowledge graph completion tasks without modifying the original language model parameters (in grey).\todo{use different colors for calibration and prompt;  add multiple arrows for calibration}}
    \label{fig:model}
\end{figure}

We reformulate KG completion as a fill-in-the-blank task. The basic idea of this task formulation is that pretrained LMs are able to answer questions formatted in cloze-style statements, and having a proper context helps to trigger LMs to produce general knowledge for the task of interest. For example, the KG completion task aims to predict the missing entity in a fact \texthrt{(Chaplin, profession, \_\_)}, which is closely related to a cloze statement. We therefore frame the KG completion as ``fill-in-the-blank'' cloze statements. In this case, ``Chaplin is a'' provides the proper context for LMs to elicit the correct answer ``screenwriter'' that is generally relevant to the task.

In more detail, a fact is in the form of \texthrt{(head, relation, tail)} or in short \texthrt{(h, r, t)}. The LM needs to predict a missing entity. A typical KG completion task provides a partial fact \texthrt{(h, r, \_\_)} and a set of candidate answers for the missing entity. To perform this task, at test time, we convert \texthrt{(h, r, t$'$)} into a cloze statement, where \texthrt{t$'$} indicates an answer candidate for filling the blank. For example, given a partial fact \texthrt{(Chaplin, profession, \_\_)}, an LM needs to fill in the blank of the cloze statement ``Charlie is a \_\_'' by providing it as the model input. In our case, a candidate answer \texthrt{(Chaplin, profession, screenwriter)} is given (e.g., ``screenwriter'' is one of the candidates), the corresponding cloze statement will turn into ``\textspt{[CLS]} Chaplin is a \textspt{[SEP]} screenwriter \textspt{[SEP]}'' (Figure~\ref{fig:model}). We use this statement as an input to a pretrained LM. \textspt{[CLS]} and \textspt{[SEP]} are special tokens of the pretrained LMs, e.g., BERT.  ``Chaplin'' is the head entity name or description. ``is a'' is relation name or description. ``screenwriter'' is the candidate tail entity name or description. Sec~\ref{sec:expset} includes resources for obtaining the entity or relation descriptions. 

\subsection{Parameter-Lite Encoder}
\label{sec:pl}
While the new formulation helps pretrained LMs to provide general knowledge about the tasks, downstream tasks often rely on task-specific or domain-specific knowledge. To adapt the general knowledge in pretrained LMs to various KG completion tasks, we introduce a parameter-lite encoder including two groups of parameters: (\expandafter{\romannumeral1}) a prompt encoder serving as the additional task-specific context in the cloze statement, and (\expandafter{\romannumeral2}) contextual calibration encoders aiming to mitigate model's bias towards general answers. The encoder is added on top of the original LM network whose parameters remain frozen during tuning.

\paragraph{Knowledge Prompt Encoder} 
Beyond general context from the task formulation, we believe that task-specific context helps better recall the knowledge of interest in pretrained LMs. For example, if we want the LM to produce the correct answer ``screenwriter'' for ``Charlie is a \_\_'', a task-specific prefix such as ``profession'' in the context will help. The LM will then assign a higher probability to ``screenwriter'' as the correct answer. In other words, we want to find a task-specific context that better steers the LM to produce task-specific predictions. Intuitively, the task-specific tokens influence the encoding of the context, thus impacting the answer predictions. However, it is non-trivial to find such task-specific tokens. For example, manually writing these tokens is not only time consuming, but also unclear whether it is optimal for our task. Therefore, we design a learnable and continuous prompt encoder.

Specifically, we use ``virtual'' prompt tokens as continuous word embeddings. As shown in Figure~\ref{fig:model}, we append these prompt tokens to different positions in the context. The embeddings of prompt tokens are randomly initialized and are updated during training. To allow more flexibility in context learning, we add a linear layer with a skip connection on top of the embedding layer to project the original token embeddings to another subspace. This projection enables learning a more tailored task-specific context that better aligns with LM's knowledge. The knowledge prompt encoder is defined in Eq.~\ref{eq:pe}.
\begin{equation}
\label{eq:pe}
    e'_i = \mathbf{W}_p e_i + b_p + e_i
\end{equation}
where $e'_i$ denotes the virtual token embedding, and $e_i$ denotes the input token embedding. $\mathbf{W}_p$ and $b_p$ are the tunable weight and bias of the prompt encoder. The knowledge prompt encoder provides task-specific context for KG completion as it is tuned on task-specific training data. 

\paragraph{Knowledge Calibration Encoder} Another main pitfall of pretrained LMs is that they tend to be biased towards common answers in their pretraining corpus. For example, the model prefers ``United States'' over ``Georgia'' for the \texthrt{birth place} of a person, which is suboptimal for KG completion. We actually view this as a shift between the pretraining distribution and the distribution of downstream tasks. 

We counteract such biases by calibrating the output distribution of pretrained LMs. Concretely, we introduce task-specific calibration parameters between Transformer layers of LMs (Figure~\ref{fig:model}) to gradually align the pretraining distribution with the downstream distribution. We choose a linear encoder with a skip connection to capture the distribution shifts, as shown in Eq.~\ref{eq:calibrate}.
\begin{equation}
\label{eq:calibrate}
    h'_i = \mathbf{W}_c h_i + b_c + h_i
\end{equation}
where $h'_i$ is the calibrated hidden state, and $h_i$ is the hidden state of a Transformer layer. $\mathbf{W}_c$ and $b_c$ are the tunable weight and bias of the \calibration.

\paragraph{Training and Inference} We keep all LM parameters fixed and only tune the parameters in the parameter-lite encoder. After formatting the KG completion tasks following our formulation, a candidate fact is in the standard sentence pair format of BERT. For example, the candidate \texthrt{(Chaplin, profession, screenwriter)} is formulated as ``\textspt{[CLS]} Chaplin is a \textspt{[SEP]} screenwriter \textspt{[SEP]}''. ``Chaplin is a'' is the first sentence as the cloze-style question, while the second sentence is ``screenwriter'' implying an answer candidate. LM then decides whether the second sentence is a correct answer to the question or not. This naturally aligns with the next sentence prediction (NSP) task of BERT, which outputs a positive label if the answer is correct; otherwise negative. Therefore, we directly utilize the next sentence prediction to perform KG completion thanks to our formulation. 

The training objective is to decide whether the second sentence is the correct next sentence to the first sentence. The small number of tunable parameters are then updated with respect to the objective. To optimize those parameters, we need both positive and negative examples. We use negative sampling~\cite{negativesample} for efficiency consideration. To be more specific, for a positive fact \texthrt{(h, r, t)}, we first corrupt its head entity with $n_\text{ns}$ random sampled entities to form negative facts, e.g., \texthrt{($\tilde{h}_i$, r, t)}. If a sampled fact is in the KG, it should be considered positive so we will re-sample it. The loss function for the head entity is defined in Eq.~\ref{eq:entl}.
\begin{equation}
\begin{aligned}
    L_\texthrt{h} = - &\log\Pr(1|h, r, t) \\
    - &\sum_{i}^{n_\text{ns}} \mathbb{E}_{\tilde{h}_i\sim E \backslash \{h\}}\log \Pr(0 | \tilde{h}_i, r, t)
\end{aligned}
\label{eq:entl}
\end{equation}
where $\Pr(\cdot|h, r, t)$ is the output probability of the BERT NSP classifier.

For each fact, the losses for its relation $L_\texthrt{r}$ and tail entity $L_\texthrt{t}$ are similarly defined. There are $3*n_\text{ns}$ negative facts in total for each fact. Similar to the negative facts for its head entity (e.g., \texthrt{($\tilde{h}_i$, r, t)}), we have the negative facts for its relation (e.g., \texthrt{(h, $\tilde{r}_i$, t)}), and its tail entity (e.g., \texthrt{(h, r, $\tilde{t}_i$)}) respectively. The joint loss function is the sum of the three components as defined in Eq.~\ref{eq:ns}.
\begin{equation}\label{eq:ns}
    L=\sum_{{(h, r, t)} \in G} (L_\texthrt{h} + L_\texthrt{r} + L_\texthrt{t})
\end{equation}
where $G$ is a collection of all KG facts.

\section{Experiments}
In this section, we evaluate the parameter-lite transfer ability of \method\ on two KG completion tasks: triplet classification and link prediction. The details of the experimental setup, datasets, and comparison methods are described in Appendix~\ref{appendix:exp}.
\subsection{Experimental Setup}
\label{sec:expset}
\paragraph{Datasets} 
We conduct the experiments on five datasets: WN11~\cite{KB_NL_1} and FB13~\cite{KB_NL_1} for triplet classification, and FB15k-237~\cite{toutanova_observed_2015}, WN18RR~\cite{ConvE} and UMLS~\cite{ConvE} for link prediction. A detailed description for these datasets is in Appendix~\ref{app:model}\todo{add}. Table \ref{tab:statistics} summarizes the statistics of the datasets.

\begin{table}[htbp]
\centering
\resizebox{0.9\linewidth}{!}
  {
    \begin{tabular}{l|c|c|c|c|c}
        \toprule
        {\bf Dataset} & $\#$ {\bf Entity} & $\#$ {\bf Relation} & $\#$ {\bf Train} & $\#$ {\bf Dev} & $\#$ {\bf Test}\\
        \hline
        FB15k-237 &  14,541 & 237 & 272,115 & 17,535 & 20,466 \\
        \hline
        WN18RR & 40,943 & 11 & 86,835 & 3,034 & 3,134 \\
        \hline
        UMLS & 135 & 46 & 5,216 & 652 & 661 \\
        \hline
        FB13 & 75,043 & 13 & 316,232 & 5,908 & 23,733 \\
        \hline
        WN11 & 38,696 & 11 & 112,581 & 2,609 &  10,544 \\
        \bottomrule
    \end{tabular}}
    \caption{Dataset statistics.}
    \label{tab:statistics}
\end{table}

\jianhao{categorize using task-specific model, general model. done}
\paragraph{Comparison Methods}
We compare \method\ with the following KG completion models.
    {\sl (\expandafter{\romannumeral1}) task-specific models (designed for KG completion)}: TransE~\cite{TransE}, 
    TransH~\cite{TransH},
    TransR~\cite{TransR}, 
    TransD~\cite{ji_knowledge_2015},
    DistMult~\cite{DistMult}, 
    TransG~\cite{xiao_transg_2016}, TranSparse~\cite{ji_knowledge_2016},  ComplEx~\cite{trouillon_complex_2016}, R-GCN~\cite{schlichtkrull_modeling_2018},
    ConvE~\cite{ConvE}, 
    ConvKB~\cite{ConvKB}, 
    DistMult-HRS~\cite{zhang_knowledge_2018},
    RotatE~\cite{RotateE},     
    REFE~\cite{chami-etal-2020-low},
    HAKE~\cite{zhang_learning_2019},
    and ComplEx-DURA~\cite{NEURIPS2020_f6185f0e},
    NTN~\cite{KB_NL_1},
    DOLORES~\cite{wang_dolores:_2018}, 
    KBGAT~\cite{nathani-etal-2019-learning},
    and GAATs~\cite{wang_knowledge_2020},
    TEKE~\cite{wang_text-enhanced_2016},
    stAR~\cite{wang_structure-augmented_2021};
    and (\expandafter{\romannumeral2}) a {\sl general model} KG-BERT~\cite{KGBERT}. It utilizes finetuning to transfer LMs for KG completion, and is task agnostic.

\subsection{Main Results}
\paragraph{Triplet Classification}
Triplet classification is a binary classification task to predict whether a given fact \texthrt{(h, r, t}) is correct or not. For each fact, we prepare the input following Sec.~\ref{sec:zs} (Figure \ref{fig:model}) and feed the input into the model. The prediction score is the output probability of the NSP classifier. If the score is above a threshold, the fact is predicted as positive, otherwise negative. We tune the threshold on dev sets and report the accuracy on test data. The results are summarized in Table \ref{tab:tc_results}. 

\begin{table}[!htb]
\centering
\resizebox{\linewidth}{!}
  {
    \begin{tabular}{l|cc|c}
       \toprule
  {{\bf Method}} & {\bf WN11} & {\bf FB13} & {\bf Avg} \\ 
    \hline
    \multicolumn{4}{l}{{\bf Task-specific models}} \\
    \hline
    {NTN~\cite{KB_NL_1}}& 86.2 & 90.0 & 88.1 \\
    {TransE~\cite{TransE}} & 75.9  & 81.5  & 78.7 \\
    {TransH~\cite{TransH}} & 78.8 & 83.3 & 81.1 \\
   {TransR~\cite{TransR}} & 85.9 &82.5  & 84.2 \\
    {TransD~\cite{ji_knowledge_2015}} &86.4  & 89.1 & 87.8 \\
    {TEKE~\cite{wang_text-enhanced_2016}} &86.1  &84.2  & 85.2  \\
    {TransG~\cite{xiao_transg_2016}} &87.4  &87.3  &87.4  \\
    {TranSparse-S~\cite{ji_knowledge_2016}} &86.4  &88.2  &87.3  \\
    {DistMult~\cite{DistMult}} &87.1  &86.2  &86.7  \\
    {DistMult-HRS~\cite{NEURIPS2020_f6185f0e}} &88.9  &89.0  &89.0  \\
    {AATE~\cite{KB_NL_6}} &88.0  &87.2  &87.6  \\
    {ConvKB~\cite{ConvKB}} &87.6  &88.8  &88.2  \\
    {DOLORES~\cite{wang_dolores:_2018}} &87.5  & 89.3   &88.4  \\
    \hline
    \hline
    \multicolumn{4}{l}{{\bf General models}} \\
    \hline
    {KG-BERT~\cite{KGBERT}} & 93.5  & 90.4   & 91.9  \\
    \hdashline
    \method$_{\rm BASE}$ (\electricblue{\small ours})& 93.3 & 91.3 & 92.3 \\
    \method$_{\rm LARGE}$  (\electricblue{\small ours}) &  93.8 &  91.7 &  92.8 \\
    \hline
    \end{tabular}}
    \caption{Triplet classification accuracy. Task-specific models are designed for knowledge graph completion, while general models are task agnostic.}
    \label{tab:tc_results}
\end{table}

\todo{check the accuracy of the numbers, e.g., improvements (i fixed some)}
\method$_{\rm BASE}$ outperforms all task-specific models, and achieves competitive or better performance compared to the finetuning method KG-BERT. \methodLarge\ gains further improvement. \method$_{\rm BASE}$ outperforms the best task-specific model by 4.4\% on WN11 and 1.3\% on FB13. It outperforms the finetuning method by 0.4\% on average (with a 0.9\% improvement on FB13). It is slightly worse than the finetuning model on WN11. Compared to finetuning, \methodLarge\ gains 0.3\% and 1.3\% improvements on WN11 and FB13 respectively. These results suggest that \method\ is able to transfer knowledge in pretrained LMs for KG completion tasks. Importantly, it is able to outperform the transfer learning performance of finetuning with much fewer parameters.

\paragraph{Link Prediction}
Link prediction aims to predict a missing entity given relation and the other entity. It is a ranking problem where we are asked to rank all candidate entities and select the top answer to complete the missing part. For each fact \texthrt{(h, r, t)}, we corrupt it by replacing either its head or tail entity with every other entity to form the candidate set. We follow \citet{TransE} to use a filtered setting, i.e., all facts that appear in either train, dev or test data are removed, and use the remaining facts as the candidate set. Similar to triplet classification, each candidate fact is fed into \method\ and the associated score is the output probability of the NSP classifier. We rank all candidates according to these scores. Two standard metrics are used for evaluation: Mean Rank (MR) and Hits@10 (the proportion of the correct entity ranked in the top 10). A lower MR is better while a higher Hits@10 is better. 

\begin{table*}[!htb]
\centering
\resizebox{0.8\linewidth}{!}
  {
    \begin{tabular}{l|cc|cc|cc}
    \toprule
   \multirow{2}{*}{{\bf Method}}  & \multicolumn{2}{c|}{{\bf FB15k-237}}  &\multicolumn{2}{c|}{{\bf WN18RR}} &  \multicolumn{2}{c}{{\bf UMLS}}\\
     &{\bf Hits@10} & {\bf MR} &  {\bf Hits@10} & {\bf MR} &{\bf Hits@10} & {\bf MR}  \\ 
    \hline
   \multicolumn{7}{l}{{\bf Task-specific models}} \\
    \hline
    TransE~\cite{TransE} & 0.465 & 357 &  0.501 & 3384 & 0.989 & 1.84 \\
    DistMult~\cite{DistMult}& 0.419 & 254 &  0.49 & 5110 &  0.846 & 5.52 \\
    ComplEx~\cite{trouillon_complex_2016}& 0.428 & 339 &  0.51 & 5261 & 0.967 & 2.59\\
    ConvE~\cite{ConvE}& 0.501 & 244 & 0.52 & 4187 & 0.990&1.51 \\
    RotatE~\cite{RotateE} & 0.533 & 177 & 0.571 & 3340 & -&- \\
    REFE~\cite{chami-etal-2020-low}& 0.541 & - & 0.561 & - & - &- \\
    HAKE~\cite{zhang_learning_2019}& 0.542 & -  & 0.582 & -  & - & - \\ 
    KBGAT~\cite{nathani-etal-2019-learning}&  0.626 & 210 &0.581 & 1940 &  - & - \\ 
    GAATs~\cite{wang_knowledge_2020}&  0.650 & 187 & 0.604 & 1270 & - & -\\ 
   StAR~\cite{wang_structure-augmented_2021}& 0.562 &   117 &   0.732 &   46 &  0.991  &1.49 \\ 
    ComplEx-DURA~\cite{NEURIPS2020_f6185f0e}& 0.560 & - & 0.571 & -  & - & -\\ 
    \hline
    \hline
    \multicolumn{7}{l}{{\bf General models}} \\
    \hline
    KG-BERT~\cite{KGBERT}& 0.420 & 153 &  0.524 & 97 & 0.990 &   1.47\\ 
    \hdashline
    \method$_{\rm BASE}$ (\electricblue{\small ours})&0.434 & 149 & 0.679 & 62 & 0.988  & 1.65\\
    \methodLarge (\electricblue{\small ours})&0.444 & 144 & 0.693 & 61 & 0.990 & 1.57\\
    \hline
    \end{tabular}}
    \caption{Link prediction results. Task-specific models are designed for knowledge graph completion, while general models are task agnostic.}
    \label{tab:lp_results}
\end{table*}


The evaluation results of link prediction are shown in Table \ref{tab:lp_results}. \method $_{\rm BASE}$ achieves competitive or better performance than the finetuning approach. \methodLarge\ performs better. In particular, \method $_{\rm BASE}$ outperforms KG-BERT by 1.4\% in Hits@10 and 4 units in MR on FB15k-237; and 15.5\% in Hits@10 and 35 units in MR on WN18RR. \methodLarge\ outperforms \method $_{\rm BASE}$ by 1\% in Hits@10, 5 units in MR on FB15k-237; and 1.4\% in Hits@10 and 1 unit in MR on WN18RR. On UMLS, the finetuning model outperforms \method $_{\rm BASE}$ by a small margin. This is because pretrained LMs contain less medical knowledge due to a lack of medical corpus during pretraining. As a result, finetuning has the advantage over our approach on UMLS. The state-of-the-art task-specific model performs better than \method. This is mainly because they leverage the structure information of KGs while the general models do not.

\subsection{Ablation Study}
To better understand \method, we further conduct an ablation study on WN11 to show the effectiveness of different components. Specifically, we evaluate \method $_{\rm BASE}$ without \prompt\ (denoted as ``w/o Prompt'') or \calibration\ (denoted as ``w/o Calibration''). We also remove the entire parameter-lite encoder (denoted as ``w/o Encoder''). Note this will make \method\ a zero-shot model since there are no tunable parameters. For comparison, we also test BERT$_{\rm BASE}$ under the finetuning setting, where we do not add any new parameters and directly finetune BERT for triplet classification with our formulation. The results are shown in Table \ref{tab:ablation}.

\begin{table}[!htb]
\centering
\resizebox{0.75\linewidth}{!}
  {
    \begin{tabular}{l|c}
       \toprule
   {\bf Method} & {\bf WN11}  \\ 
    \hline
    \method $_{\rm BASE}$  & 93.3 \\
    w/o Prompt & 91.7 \\
    w/o Calibration$_{\rm middle}$ & 92.2 \\
    w/o Calibration$_{\rm last}$  & 93.0 \\
    w/o Calibration$_{\rm both}$ & 89.3 \\
    w/o Encoder & 73.7 \\
    \hdashline
    Finetuning & 93.2 \\
    \bottomrule
    \end{tabular}}
    \caption{Ablation study on WN11. We remove \prompt, or \calibration, or the entire parameter-lite encoder.\jianhao{check consistency}}
    \label{tab:ablation}
\end{table}

We have the following observations: (\expandafter{\romannumeral1}) all components have a positive effect on the final performance. The \prompt\ brings the most improvement which is 1.6\%. The \calibration\ at the middle layer brings a 1.1\% improvement, and that at the last layer brings a 0.3\% improvement. The results indicate that it is more important to recall and prepare the knowledge in earlier layers for the task of interest. (\expandafter{\romannumeral2}) Removing both \calibration s results in the worst accuracy. The \calibration s are important for knowledge transfer. (\expandafter{\romannumeral3}) \method $_{\rm BASE}$ outperforms finetuning all parameters, which suggests that \method\ is an effective way to adapt pretrained LMs for KG completion since it requires far less computation and storage. (\expandafter{\romannumeral4}) Furthermore, we can see that without the entire \pl\ encoder, our model still achieves promising results. On WN11, it achieves 73.7\% accuracy, which is approximately 1.5 times the accuracy of random guesses (50\%). This shows the effectiveness of our task formulation. Formulating KG completion as a ``fill-in-the-blank'' task triggers the knowledge that an LM learned during pretraining. This enables our efficient transfer algorithm.

\subsection{Parameter Efficiency Analysis}
\begin{figure}[!ht]
    \centering
    \includegraphics[width=0.4\textwidth]{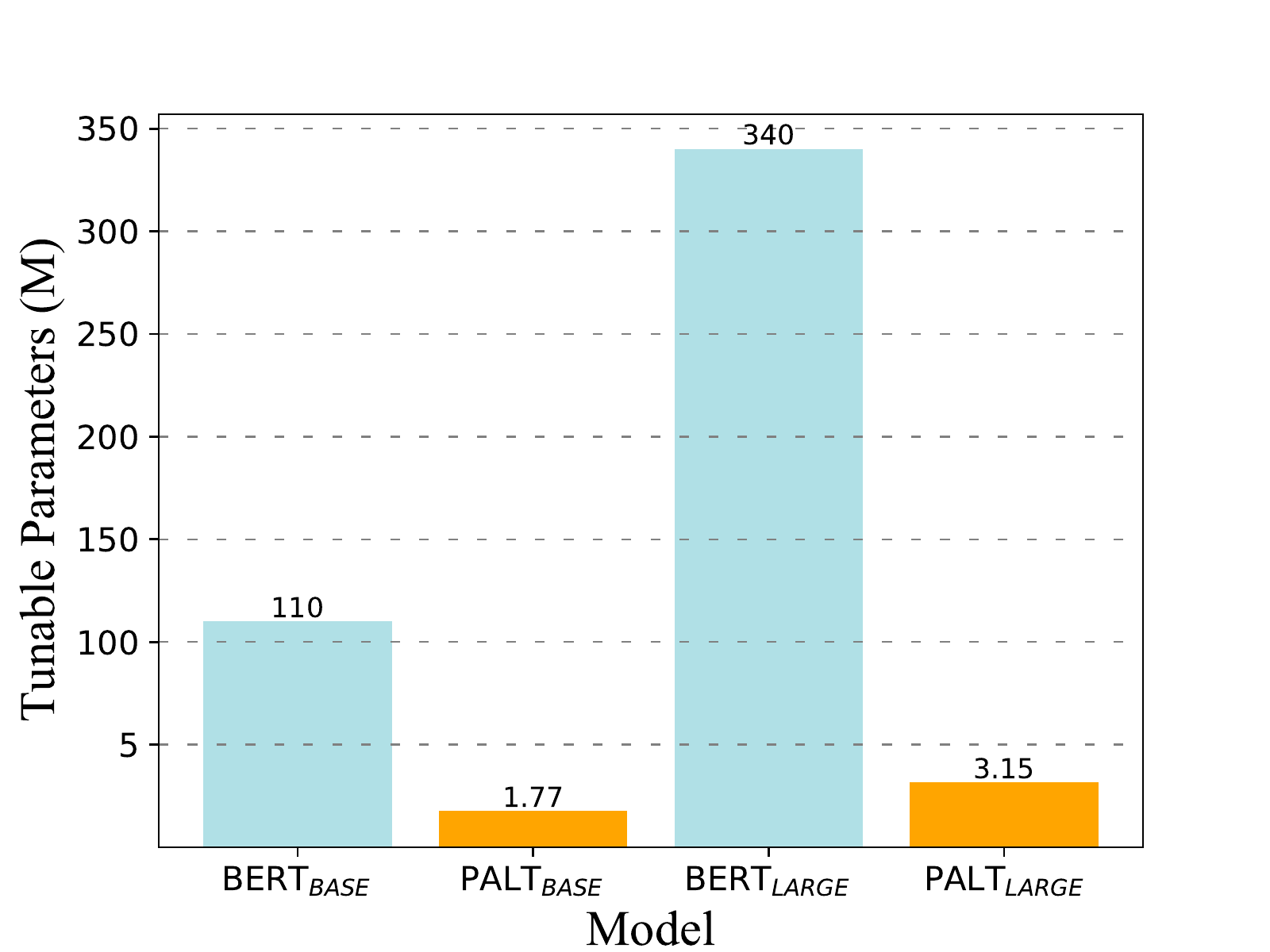}
    \caption{Compare the number of tunable parameters of \method\ and BERT.}
    \label{fig:parameter}
\end{figure}

The advantage of \method\ is that only a small amount of newly added parameters are tuned while all LM parameters are fixed. This brings two benefits: space-efficient model storage and efficient computation. Here we compare the numbers of tunable parameters of \method\ and BERT, and the detailed calculation of \method\ is presented in Appendix~\ref{appendix:param}\todo{add citation}. BERT$_{\rm BASE}$ has 110M tunable parameters, while \method $_{\rm BASE}$ has 1.77M. \methodLarge\ has 3.15M tunable parameters, while BERT$_{\rm LARGE}$ has 340M. \method\ $_{\rm BASE}$ only has 1.6\% tunable parameters of BERT$_{\rm BASE}$, and \method\ $_{\rm LARGE}$ has 0.9\% of BERT$_{\rm LARGE}$. We show the numbers of tunable parameters of different models in Figure \ref{fig:parameter}.

\subsection{Case Study}
\begin{figure*}[h]
    \centering
    \subfigure[The first layer.]{
    \begin{minipage}[t]{0.3\linewidth}
    \includegraphics[width=\linewidth]{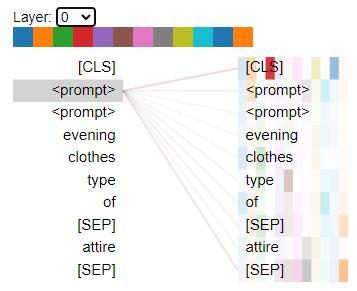}
    \end{minipage}
    }
    \subfigure[The middle layer.]{
    \begin{minipage}[t]{0.3\linewidth}
    \includegraphics[width=\linewidth]{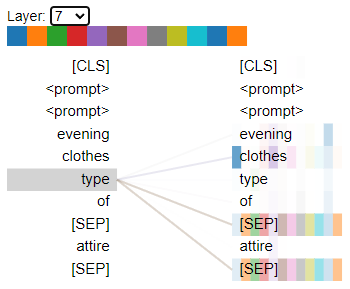}
    \end{minipage}
    }
    \subfigure[The last layer.]{
    \begin{minipage}[t]{0.3\linewidth}
    \includegraphics[width=\linewidth]{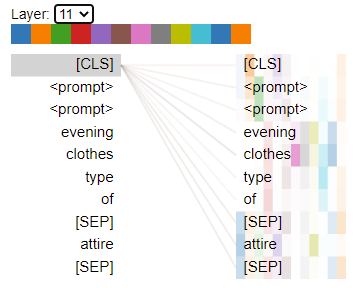}
    \end{minipage}
    }
    \caption{Visualization of attention weights of different Transformer layers of \method. The 0th layer is the first attention layer. The 7th layer is the attention layer after our middle calibration encoder. The 11th layer is the last attention layer. Different color represents different attention heads. The darker the color is, the larger the attention score.}
    \label{fig:visualization}
\end{figure*}

In this section, we perform a case study analysis to illustrate why \method\ performs well. We use BertViz~\cite{vig_multiscale_2019} to visualize the attention weights of \method . We take an example of a positive fact \texthrt{(h, r, t)}:
$h =$ \texthrt{``evening clothes''}, $r =$ \texthrt{``type of''} and $t =$ \texthrt{``attire''} and show the attention weights of the first, the middle and the last attention layers in Figure \ref{fig:visualization}. In the first layer, the prompt token attends to all tokens, indicating that it helps to recall general knowledge. In the middle layer, the attention weights concentrate on the most relevant parts of the tokens. Specifically, the attention weight between \texthrt{``type''} and \texthrt{``clothes''} is large. The \texthrt{``type''} token also pays attention to \textspt{[SEP]} token. It is mainly because the \textspt{[SEP]} token marks the boundary of two sentences and the pretrained LM uses it as an aggregation representation of each sentence. In the last layer, different heads of \textspt{[CLS]} focus on different parts of the text. For example, the first head (in blue) attends to the tail entity. The seventh head (in pink) attends to the head and relation. The third head (in green) attends to prompt tokens. This shows that \textspt{[CLS]} gathers task-specific knowledge for the NSP classifier.

In Table \ref{tab:good_case_fb13}, we give some examples of FB13 that are improved by \method\ compared to the finetuning approach (i.e., facts that are correctly predicted by \method\ while KG-BERT fails). We further show the comparison between attention weights of \method\ and the finetuning approach in Appendix~\ref{app:attention}\todo{add}.

\begin{table}[!htbp]
    \centering
    \resizebox{0.98\linewidth}{!}{
    \begin{tabular}{cccc}
     \toprule
    Head & Relation & Tail & Label \\
    \midrule
    tetsuzan nagata & cause of death & murder & \correctmark \\
    charles eliot & gender & male & \correctmark \\
    bill burrud & institution & harvard university & \correctmark \\
    lothar rendulic & nationality & germany & \correctmark \\
    thomas abbt & profession & philosopher & \correctmark \\
    samuel richardson & profession & priest & \wrongmark \\
    nathaniel wallich & ethnicity & white british & \wrongmark \\
    fu biao & gender & female & \wrongmark \\
    alan turing & institution & harvard law school & \wrongmark \\
    alan turing & ethnicity & israelis & \wrongmark \\
    \bottomrule
    \end{tabular}}
    \caption{Samples of \method's correct predictions on FB13, where the finetuning method~\cite{KGBERT} outputs wrong predictions. Label \correctmark\ means gold positive fact and \wrongmark\ indicates gold negative fact.}
    \label{tab:good_case_fb13}
\end{table}

\subsection{Error Analysis}
\begin{figure*}[htbp]
    \centering
    \subfigure[\texthrt{Has Instance.}]{
    \begin{minipage}[t]{0.3\linewidth}
    \includegraphics[width=\linewidth]{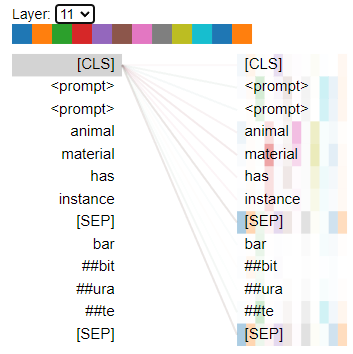}
    \end{minipage}
    }
    \subfigure[\texthrt{Domain Topic.}]{
    \begin{minipage}[t]{0.3\linewidth}
    \includegraphics[width=\linewidth]{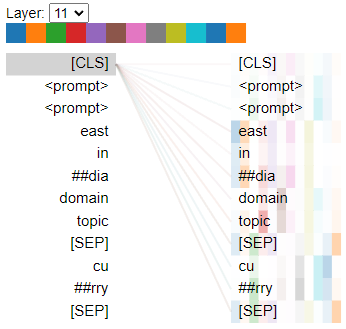}
    \end{minipage}
    }
    \subfigure[\texthrt{Subordinate Instance.}]{
    \begin{minipage}[t]{0.3\linewidth}
    \includegraphics[width=\linewidth]{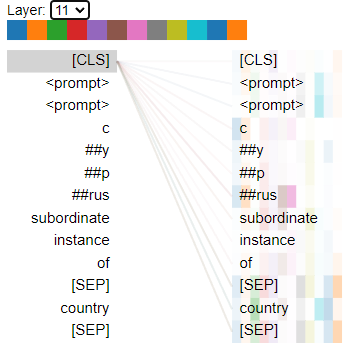}
    \end{minipage}
    }
    \caption{The attention weights of the last layer of \method\ on three error cases involving different relations.}
    \label{fig:error}
\end{figure*}

We analyze the errors made by \method\ in this section. Here we focus on analyzing relations with the highest and lowest error rates. The detailed error rate statistics are shown in Appendix~\ref{app:error}\todo{add}. Most of \method\ errors are due to \texthrt{``domain''} relations, with an error rate of 14.7\% for the relation \texthrt{``domain topic''} and 10.5\% for \texthrt{``domain region''}. The reason is that we find the relations of the ``domain'' are not well defined, and the boundary between relations can be unclear. For the relation \texthrt{``subordinate instance of''}, \method\ performs the best with an error rate of 2.6\%, since it is more related to semantic information. We further analyze the attention weights of some error cases in Figure \ref{fig:error}. For the first case, the \textspt{[CLS]} token attends mainly to the head and relation tokens but little to the tail entity. This is because \texthrt{``barbiturate''} is a rare entity and the LM does not capture much knowledge for it during pretraining. \method\ fails on the second case mainly because \texthrt{``domain topic''} covers a wide range of concepts. This results in a uniform distribution of attention so it is difficult to make a correct prediction. For the last case, \textspt{[CLS]} attends to both the head and tail entities but little to the relations. This leads to the error. We notice that most entities are segmented into sub-words based on BERT's tokenizer. This may result in a poor understanding of entities. We believe other pretraining paradigms like span-masking~\cite{joshi-etal-2020-spanbert} will help and leave it as future work. 

\section{Related Work}
Pretrained LMs~\cite{BERT, liu_roberta:_2019, radford2019language,brown_language_2020, lewis_bart:_2020, raffel_exploring_2020,wang-etal-2022-deepstruct} have achieved state-of-the-art results in many NLP tasks~\cite{wang_glue:_2018, wang_superglue:_2019}. Some work also uses pretrained LMs for knowledge-driven tasks~\cite{KGBERT,LM4KG,inspecting} by finetuning them on downstream tasks, which has been the de facto method to achieve superior results. However, finetuning modifies all parameters, which requires a large amount of computation and storage resources. Recently, prompt-tuning~\cite{shin-etal-2020-autoprompt, liu_gpt_2021}, adaptors~\cite{pmlr-v97-houlsby19a, k-adapter, newman_p-adapters:_2021}, and factual probing~\cite{he2021empirical, LAMA, wang2020language, wang2021zero} are developed to transfer pretrained LMs to downstream tasks and show improvements on many NLP tasks. \citet{few_learn} focus on prompt based finetuning, while \citet{lm_know} follow standard prompt formulation for a single token. 

Compared to the existing methods, there are several distinctive features of \method. First, instead of using the output of a single \textspt{[MASK]} token, we leverage the next sentence prediction, which allows the answers with arbitrary lengths. Second, we automatically acquire the template using the natural language descriptions of the relations available in the downstream KGs. Besides, we also use the corresponding entity descriptions in the cloze statement, providing richer context. Third, our method differs from prompt-tuning, which only inserts virtual tokens into the input. Fourth, in contrast to the empirical calibration procedures that are highly customized for each task, our method automatically learns a few calibration parameters for each task. Overall, compared to previous methods, our approach is lightweight. The parameter-lite encoder is unique and particularly useful for more NLP tasks.

Traditional KG completion methods mainly rely on graph structure information. They embed entities and relations into a continuous vector space and learn a score function based on these embeddings for triplets~\cite{TransE, TransH, TransR, Rescal, Tucker, ConvE, ConvKB, cai_kbgan:_2018}. Unlike \method, these methods treat entities and relations as unique identifiers and ignore their semantic meaning. Another line of research leverages text descriptions of entities and relations for KG completion. For example, KG-BERT~\cite{KGBERT} concatenates the text description of entities and relations into a sequence and feeds it into BERT, and finetunes the task-specific models. LASS~\cite{lass} further uses both text and structure information to solve different KG completion tasks under a unified LM finetuning framework. By contrast, we present an alternative to finetuning for KG completion, and our method unifies different tasks in the same model architecture.
\section{Conclusion}
We propose \method, a parameter-lite transfer of pretrained language models (LM) for knowledge graph completion. To efficiently elicit general knowledge of LMs learned about the task during pretraining, we reformulate KG completion as a ``fill-in-the-blank'' task. We then develop a parameter-lite encoder including two groups of parameters. First, it contains a knowledge prompt encoder consisting of learnable continuous prompt tokens to better recall task-specific knowledge from pretrained LMs. Second, it calibrates pretrained LMs representations and outputs for KG completion via two \calibration s. As a result, our method achieves competitive or even better results than finetuning with far fewer tunable parameters. Both the task formulation and parameter-lite encoder can be inspiring for a wide range of knowledge-intensive tasks and deep LMs. We hope this research can foster future research along the parameter-lite knowledge transfer direction in NLP.

\section{Limitations}
As for the limitations of our method, the input is constructed based on the natural language descriptions of the entities and relations, and such descriptions may need additional efforts to obtain in different application scenarios. Although our method achieves competitive results in the medical domain (UMLS), the main finding of our study is that our method is more capable of transferring general knowledge in LMs to KG completion tasks. We welcome more studies on strengthening its performance in specific domains, e.g., using domain-specific LMs for a particular domain (e.g., BioBERT~\cite{biobert} for the medical domain). Finally, our method shares some common limitations with most deep learning approaches. For example, the decisions are not easy to interpret, and the predictions can retain the biases of the training data.

\section{Ethical Considerations}
We hereby acknowledge that all of the co-authors of this work are aware of the provided \textit{ACM Code of Ethics} and honor the code of conduct. The followings give the aspects of both our ethical considerations and our potential impacts to the community.
This work uses pretrained LMs for KG completion. We develop an encoder especially the knowledge calibration encoder to mitigate the potential knowledge biases in pretrained LMs. The risks and potential misuse of pretrained LMs are discussed in \cite{brown_language_2020}. There are potential undesirable biases in the datasets, such as unfaithful descriptions from Wikipedia. We do not anticipate the production of harmful outputs after using our model, especially towards vulnerable populations. 

\section{Environmental Considerations}
We build \method\ based on pretrained BERT$_{\rm BASE}$ and BERT$_{\rm LARGE}$. According to the estimation in \cite{strubell-etal-2019-energy}, pretraining a base model costs 1,507 kWh$\cdot$PUE and emits 1,438 lb $CO_2$, while pretraining a large model requires 4 times the resources of a base model. Our methods only tune 1\% parameters with fewer than 1\% gradient-steps of the number of steps of pretraining. Therefore, our energy cost and $CO_2$ emissions are relatively small. 
\section*{Acknowledgements}
We would like to thank the anonymous reviewers for their suggestions and comments.
This paper is partially supported by National Key Research and Development Program of China with Grant No. 2018AAA0101902 and the National Natural Science Foundation of China (NSFC Grant Numbers 62106008 and 62276002).
This material is in part based upon work supported by Berkeley DeepDrive and Berkeley Artificial Intelligence Research.
The research was also supported in part by US DARPA KAIROS Program No. FA8750-19-2-1004 and INCAS Program No. HR001121C0165, National Science Foundation IIS-19-56151, IIS-17-41317, and IIS 17-04532.

\bibliography{anthology}
\bibliographystyle{acl_natbib}

\appendix
\section{Experimental Setup Details}
\label{appendix:exp}
We describe additional details of our experimental setup including implementation, datasets and comparison methods in this section.

\subsection{Implementation Details}
We implement our algorithm using the Hugging Face Transformers package. We optimize \method\ with AdamW~\cite{loshchilov_decoupled_2018}. The hyper-parameters are set as follows. We use 8 GPUs and set the batch size to 32 per GPU, and set the learning rate to [$1.5*10^{-4}$, $1.5*10^{-5}$, $1*10^{-4}$, $1*10^{-5}$] for WN11, FB13, FB15k-237, WN18RR, respectively. We set the warm-up ratio to 0.1 and set weight decay as 0.01. The number of training epochs is 10 for link prediction and 40 for triplet classification. For link prediction, we sample 5 negative samples for the head entity, relation and tail entity, resulting in 15 negative triplets in total for each sample. And for triplet classification, we only sample one negative sample for each entity. Note that the negative samples here are used for training (Eq. \ref{eq:ns}), which are different from the candidate sets for link prediction during evaluation. 
We adopt grid search to tune the hyper-parameters on the dev set. For learning rates, we search from 1e-5 to 5e-4 with an interval of 5e-6. For the number of negative examples, we test values in \{1, 5, 10\}. For the remaining hyper-parameters, we generally follow BERT’s setup. 

For model inputs, we use synset definitions as entity descriptions for WN18RR, and descriptions produced by \citet{KB_NL_3} for FB15k-237. For FB13, we use entity descriptions in Wikipedia. We use entity names for WN11 and UMLS. For all datasets, we use relation names as relation descriptions. 

For \method\ architecture, we insert two \calibration s to the middle layer and last layer of BERT. This applies to both \method$_{\rm BASE}$ and \methodLarge. For \prompt, we add it to the input layer. In particular, 10 prompt tokens are added at 3 different positions for \method$_{\rm BASE}$ on all datasets except for WN11. 2 prompt tokens are added at the beginning of WN11. This is because the entity description of WN11 is short. For \methodLarge, we add 2 prompt tokens at the beginning.

\subsection{Datasets}
We introduce the link prediction and triplet classification datasets below.

\subsubsection{Link Prediction}
\begin{itemize}[leftmargin=*]
\item {\bf FB15k-237}. Freebase is a large collaborative KG consisting of data composed mainly by its community members. It is an online collection of structured data harvested from many sources, including individual and user-submitted wiki contributions \cite{freebase}. FB15k is a selected subset of Freebase that consists of 14,951 entities and 1,345 relationships \cite{TransE}. FB15K-237 is a variant of FB15K where inverse relations and redundant relations are removed, resulting in 237 relations \cite{text_joint_kb}.

\item {\bf WN18RR}. WordNet is a lexical database of semantic relations between words in English. WN18 \cite{TransE} is a subset of WordNet which
consists of 18 relations and 40,943 entities. WN18RR is created to ensure that the evaluation dataset does not have inverse relations to prevent test leakage \cite{ConvE}.

\item {\bf UMLS}. UMLS semantic network \cite{UMLS} is an upper-level ontology of the Unified Medical Language System. The semantic network, through its 135 semantic types, provides a consistent categorization of all concepts represented in the UMLS. The 46 links between the semantic types provide the structure for the network and represent important relationships in the biomedical domain.

\comm{
\item {\bf YAGO3-10}. Yet Another Great Ontology (YAGO) is a KG that augments WordNet with common knowledge facts extracted from Wikipedia, converting WordNet from a primarily linguistic resource to a common KG \cite{yago}. YAGO3-10 is a benchmark dataset for KG completion. It is a subset of YAGO3 (which itself is an extension of YAGO) that contains entities associated with at least ten different relations. Table~\ref{tab:yago} shows the statistics of YAGO3-10 dataset.
}
\end{itemize}

\comm{
\begin{table}[htbp]
    \centering
    \begin{tabular}{ccccc}
    \toprule
        \# {\bf Entity} & \# {\bf Relation} & \# {\bf Train} & \# {\bf Dev} & \# {\bf Test}  \\
        \midrule
        123,182  & 37 & 1,079,040 & 5,000 & 5,000 \\
        \bottomrule
    \end{tabular}
    \caption{Statistics of YAGO3-10 dataset.}
    \label{tab:yago}
\end{table}
}

\subsubsection{Triplet Classification}
\begin{itemize}[leftmargin=*]
\item {\bf WN11 and FB13} are subsets of WordNet and Freebase respectively for triplet classification, where the \citet{KB_NL_1} randomly switch entities from correct testing triplets resulting in a total of doubling the number of test triplets with an equal number of positive and negative examples.
\end{itemize}

\subsection{Comparison Methods}
\label{app:model}
We compare \method\ to three types of KG completion methods: shallow structure embedding, deep structure embedding, and language semantic embedding.\footnote{We refer the readers to \cite{ji2021survey} for a more comprehensive review of the KG completion methods.}

\subsubsection{Shallow Structure Embedding}
TransE~\cite{TransE}, TransH~\cite{TransH}, TransR~\cite{TransR}, TransD~\cite{ji_knowledge_2015}, TransG~\cite{xiao_transg_2016}, TranSparse-S~\cite{ji_knowledge_2016}, DistMult~\cite{DistMult}, ConvKB~\cite{ConvKB}, ComplEx~\cite{trouillon_complex_2016}, ConvE~\cite{ConvE}, RotatE~\cite{RotateE}, REFE~\cite{chami-etal-2020-low}, HAKE~\cite{zhang_learning_2019}, and ComplEx-DURA~\cite{NEURIPS2020_f6185f0e} are methods based only on the structure of the KG. DistMult-HRS~\cite{zhang_knowledge_2018} is an extension of DistMult which is combined with a three-layer hierarchical relation structure (HRS) loss. Each of these methods proposes a scoring function regarding a knowledge fact, without using the natural language descriptions or names of entities or relations. The scoring functions are shown in Table~\ref{tab:structure_methods}.

\newcommand{\norm}[1]{\left\lVert#1\right\rVert}
\begin{table*}[htbp]
\centering
\resizebox{1.0\linewidth}{!}{
\begin{tabular}{lcc}
\toprule
\textbf{Method} &  \multicolumn{2}{c}{\textbf{Score Function}} \\
\midrule
TransE & $-\norm{\textbf{h} + \textbf{r} - \textbf{t}}$ & $\textbf{h}, \textbf{r}, \textbf{t} \in \mathbb{R}^k$\\
TransH & $-\left\|\left(\mathbf{h}-\mathbf{w}_{r}^{\top} \mathbf{h} \mathbf{w}_{r}\right)+\mathbf{r}-\left(\mathbf{t}-\mathbf{w}_{r}^{\top} \mathbf{t} \mathbf{w}_{r}\right)\right\|$ & $\mathbf{h}, \mathbf{t}, \mathbf{r}, \mathbf{w}_{r} \in \mathbb{R}^{k}$\\
TransR & $-\norm{\mathbf{M}_r\mathbf{h} + \mathbf{r} - \mathbf{M}_r\mathbf{t}}$ & $\mathbf{h}, \mathbf{t}\in\mathbb{R}^k, \mathbf{M}_r\in\mathbb{R}^{k\times d}$ \\
TransD & $-\left\|\left(\mathbf{w}_{r} \mathbf{w}_{h}^{\top}+\mathbf{I}\right) \mathbf{h}+\mathbf{r}-\left(\mathbf{w}_{r} \mathbf{w}_{t}^{\top}+\mathbf{I}\right) \mathbf{t}\right\|$ & $\mathbf{h}, \mathbf{t}, \mathbf{w}_{h} \mathbf{w}_{t} \in \mathbb{R}^{k}, \mathbf{r}, \mathbf{w}_{r} \in \mathbb{R}^{d}$\\
TransG & $\sum_{i} \pi_{r}^{i} \exp \left(-\frac{\left\|\boldsymbol{\mu}_{h}+\boldsymbol{\mu}_{r}^{i}-\boldsymbol{\mu}_{t}\right\|}{\sigma_{h}^{2}+\sigma_{t}^{2}}\right)$ & $\mathbf{h} \sim \mathcal{N}\left(\boldsymbol{\mu}_{h}, \boldsymbol{\sigma}_{h}^{2} \mathbf{I}\right)$,
$\mathbf{t} \sim \mathcal{N}\left(\boldsymbol{\mu}_{t}, \Sigma_{t}\right)$,
$\boldsymbol{\mu}_{h}, \boldsymbol{\mu}_{t} \in \mathbb{R}^{k}$\\
TranSparse-S & $-\left\|\mathbf{M}_{r}\left(\theta_{r}\right) \mathbf{h}+\mathbf{r}-\mathbf{M}_{r}\left(\theta_{r}\right) \mathbf{t}\right\|_{1 / 2}^{2}$
$-\left\|\mathbf{M}_{r}^{1}\left(\theta_{r}^{1}\right) \mathbf{h}+\mathbf{r}-\mathbf{M}_{r}^{2}\left(\theta_{r}^{2}\right) \mathbf{t}\right\|_{1 / 2}^{2}$ & $\mathbf{h}, \mathbf{t} \in \mathbb{R}^{k}$, $\mathbf{r} \in \mathbb{R}^{d}, \mathbf{M}_{r}\left(\theta_{r}\right) \in \mathbb{R}^{k \times d}$, $\mathbf{M}_{r}^{1}\left(\theta_{r}^{1}\right), \mathbf{M}_{r}^{2}\left(\theta_{r}^{2}\right) \in \mathbb{R}^{k \times d}$\\
DistMult & $ \langle \textbf{r}, \textbf{h}, \textbf{t} \rangle$ & $\textbf{h}, \textbf{r}, \textbf{t} \in \mathbb{R}^k$\\
ConvKB & $\operatorname{concat}(g([\boldsymbol{h}, \boldsymbol{r}, \boldsymbol{t}] * \omega)) \mathbf{w}$ & $\textbf{h}, \textbf{r}, \textbf{t} \in \mathbb{R}^k$ \\
ComplEx & $ \Re(\langle \textbf{r}, \textbf{h}, \overline{\textbf{t}} \rangle)$ & $\textbf{h}, \textbf{r}, \textbf{t} \in \mathbb{C}^k$\\
ConvE & $ \langle \sigma(\mathrm{vec}(\sigma([ \overline{\textbf{r}}, \overline{\textbf{h}}] \ast \boldsymbol{\Omega})) \mathbf{W}), \textbf{t} \rangle$ & $\textbf{h}, \textbf{r}, \textbf{t} \in \mathbb{R}^k$\\
RotatE & $-\norm{\textbf{h} \circ \textbf{r} - \textbf{t}}^2$ & $\textbf{h}, \textbf{r}, \textbf{t} \in \mathbb{C}^k, |r_i| = 1$ \\
REFE & $-\mathrm{arctanh}(\norm{-\langle\mathbf{h}, \mathrm{Ref}(\mathbf{r})\rangle\oplus^c \mathbf{t}})$ & $\textbf{h}, \textbf{r}, \textbf{t} \in \mathbb{R}^k$\\
HAKE & $\text{RotatE}-\norm{\sin((\mathbf{h} + \mathbf{r} - \mathbf{t}) / 2)}_1$ & $\textbf{h}, \textbf{r}, \textbf{t} \in \mathbb{R}^k$ \\
ComplEx-DURA & $\text{ComplEx} - \langle \mathbf{h}, \mathbf{r}\rangle^2 - \norm{\mathbf{t}}^2$ & $\textbf{h}, \textbf{r}, \textbf{t} \in \mathbb{C}^k$ \\
\bottomrule
\end{tabular}}
\caption{The score functions $f_r(\textbf{h}, \textbf{t})$ of shallow structure embedding models for KG embedding, where $\langle \cdot \rangle$ denotes the generalized dot product, $\circ$ denotes the Hadamard product, $\sigma$ denotes activation function and $\ast$ denotes 2D convolution. $\overline{\ \cdot\ }$ denotes conjugate for complex vectors, and 2D reshaping for real vectors in the ConvE model. $\mathrm{Ref}(\theta)$ denotes the reflection matrix induced by rotation parameters $\theta$. $\oplus^c$ is Möbius addition that provides an analogue to Euclidean addition for hyperbolic space.
\label{tab:structure_methods}}
\end{table*}

\subsubsection{Deep Structure Embedding}
\begin{itemize}[leftmargin=*]
\item {\bf NTN} (Neural Tensor Network)~\cite{KB_NL_1} models entities across multiple dimensions by a bilinear tensor neural layer.

\item {\bf DOLORES}~\cite{wang_dolores:_2018} is based on bi-directional LSTMs and learns deep representations of entities and relations from constructed entity-relation chains.

\item {\bf KBGAT} proposes an attention-based feature embedding that captures both entity and relation features in any given entity’s neighborhood, and additionally encapsulates relation clusters and multi-hop relations \cite{nathani-etal-2019-learning}.

\item {\bf GAATs} integrates an attenuated attention mechanism in a graph neural network to assign different weights in different relation paths and acquire the information from the neighborhoods \cite{wang_knowledge_2020}. 
\end{itemize}

\subsubsection{Language Semantic Embedding}
\begin{itemize}[leftmargin=*]
\item {\bf TEKE}~\cite{wang_text-enhanced_2016} takes advantage of the context information in a text corpus. The textual context information is incorporated to expand the semantic structure of the KG and each relation is enabled to own different representations for different head and tail entities.

\item {\bf AATE}~\cite{KB_NL_6} is a text-enhanced KG representation learning method, which can represent a relation/entity with different representations in different facts by exploiting additional textual information.

\item {\bf KG-BERT}~\cite{KGBERT} considers facts in KG as textual sequences, where each textual sequence is a concatenation of text descriptions of the head entity, the relation, and the tail entity. Then KG-BERT treats the KG completion task as a text binary classification task, and then solves it by fine-tuning a pre-trained BERT.

\item {\bf StAR}~\cite{wang_structure-augmented_2021} partitions each fact into two asymmetric parts as in translation-based graph embedding approach, and encodes both parts into contextualized representations by a Siamese-style textual encoder (BERT or RoBERTa)~\cite{wang_structure-augmented_2021}.
\end{itemize}

\section{Number of Tunable Parameters}
\label{appendix:param}
Here we give a detailed calculation of the number of tunable parameters of \method. We denote $d_e$ as the dimension of token embeddings and $d_h$ as the hidden size in the pretrained LM, and $n_p$ as the number of prompts added in \method. The number of tunable parameters for the prompt embeddings and linear mapping weights is $n_p * d_e + d_e * d_h$, and that of \calibration\ is $2*d_h*d_h$. In total, there are $n_p*d_e + d_e * d_h + 2*d_h*d_h$ tunable parameters in \method. For BERT$_{\rm BASE}$, $d_e = d_h = 768$, and for BERT$_{\rm LARGE}$, $d_e = d_h = 1024$, and we use $n_p=2$ on WN11. As a result, \method $_{\rm BASE}$ has 1.77M tunable parameters, and \methodLarge\ has 3.15M.

\section{Case Study}
\label{app:attention}
\begin{figure}[htbp]
    \centering
    \subfigure[KG-BERT]{
    \begin{minipage}[t]{0.45\linewidth}
    \includegraphics[width=\linewidth]{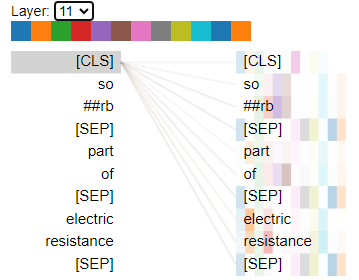}
    \end{minipage}
    }
    \subfigure[\method]{
    \begin{minipage}[t]{0.45\linewidth}
    \includegraphics[width=\linewidth]{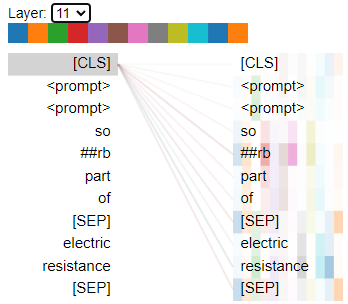}
    \end{minipage}
    }
    \caption{{\small Comparison between the attention weights of \method\ and KG-BERT. In this example, \method\ correctly predicts it as negative and KG-BERT fails.}}
    \label{fig:goodcase}
\end{figure}

To illustrate the difference between \method\ and KG-BERT, we show the attention weights of \textspt{[CLS]} in the last layer in Figure \ref{fig:goodcase}. We can see that for KG-BERT, most attention heads attend to the whole sequence, while for \method\ each head attends to a specific part of the sequence. For example, the third (colored green) head has large weights on prompts and the tail entity, and the fourth (colored red) head pays attention to the head entity and relation. This demonstrates that \method\ recalls and calibrates related knowledge in a more disentangled way than KG-BERT, and as a result, it succeeds to predict this triplet as negative.

\begin{figure*}[t]
    \centering
    \subfigure[BERT-middle]{
    \begin{minipage}[t]{0.225\linewidth}
    \includegraphics[width=\linewidth]{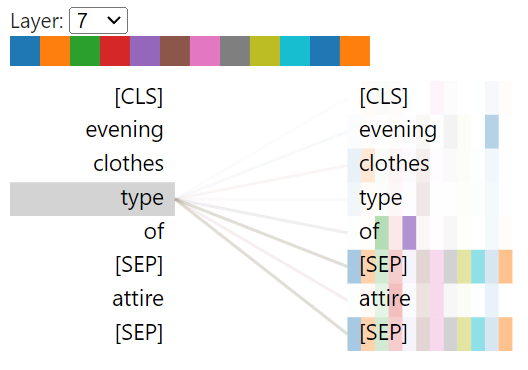}
    \end{minipage}
    }
    \subfigure[\method-middle]{
    \begin{minipage}[t]{0.225\linewidth}
    \includegraphics[width=\linewidth]{fig/layer7.png}
    \end{minipage}
    }
    \subfigure[BERT-last]{
    \begin{minipage}[t]{0.225\linewidth}
    \includegraphics[width=\linewidth]{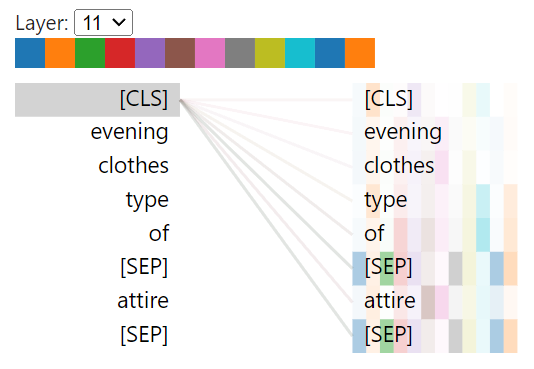}
    \end{minipage}
    }
    \subfigure[\method-last]{
    \begin{minipage}[t]{0.225\linewidth}
    \includegraphics[width=\linewidth]{fig/attention-3.png}
    \end{minipage}
    }
    \caption{{\small Attentions weights of the original BERT and \method .}}
    \label{fig:cmp_bert}
\end{figure*}

In Table \ref{tab:good_case_wn11} we demonstrate some triplets of WN11 that are correctly predicted by \method\ while KG-BERT fails. 

\begin{table}[htbp]
    \centering
    \resizebox{0.9\linewidth}{!}{
    \begin{tabular}{cccc}
     \toprule
    Head & Relation & Tail & Label \\
    \hline
    center & has instance & olfactory brain & \correctmark \\
    family graminaceae & member meronym & meadow grass & \correctmark \\
    botany & domain region & style & \correctmark \\
    end & has instance & complete & \correctmark \\
    fictionalise & type of & convert & \correctmark \\
    archaeology & domain region & unreactive & \wrongmark \\
    cognitive content & has instance & diacritic & \wrongmark \\
    anura & member meronym & kuru & \wrongmark \\
    atlantic &has part& tocantins& \wrongmark \\
    sorb &part of& electric resistance & \wrongmark \\
    \bottomrule
    \end{tabular}}
    \caption{Triplets of WN11 that are correctly predicted by \method\ while KG-BERT fails. Label \correctmark means a positive triplet and \wrongmark means negative.}
    \label{tab:good_case_wn11}
\end{table}

\section{Error Analysis}
\label{app:error}
Here we give the error rate of \method $_{\rm BASE}$ on each relation of WN11 in Table \ref{tab:error}. ``Domain topic" and ``domain region" are the two relations with the highest error rates, while ``subordinate instance of" has the lowest error rate. 

\begin{table}[htbp]
    \centering
    \begin{tabular}{c|c}
     \toprule
    Relation & Error Rate(\%) \\
    \hline
    domain topic & 14.7 \\
    domain region & 10.5 \\
    has instance & 8.3 \\
    member meronym & 8.1 \\
    synset domain topic & 7.4 \\
    similar to & 7.1 \\
    has part & 7.0 \\
    part of & 5.7 \\
    type of & 5.3 \\
    member holonym & 3.9 \\
    subordinate instance of & 2.6 \\
    \bottomrule
    \end{tabular}
    \caption{{\small The error rates of triplet classification on different relations.}}
    \label{tab:error}
\end{table}

\section{Prompt Analysis}
\begin{figure}[ht]
    \centering
    \includegraphics[width=0.4\textwidth]{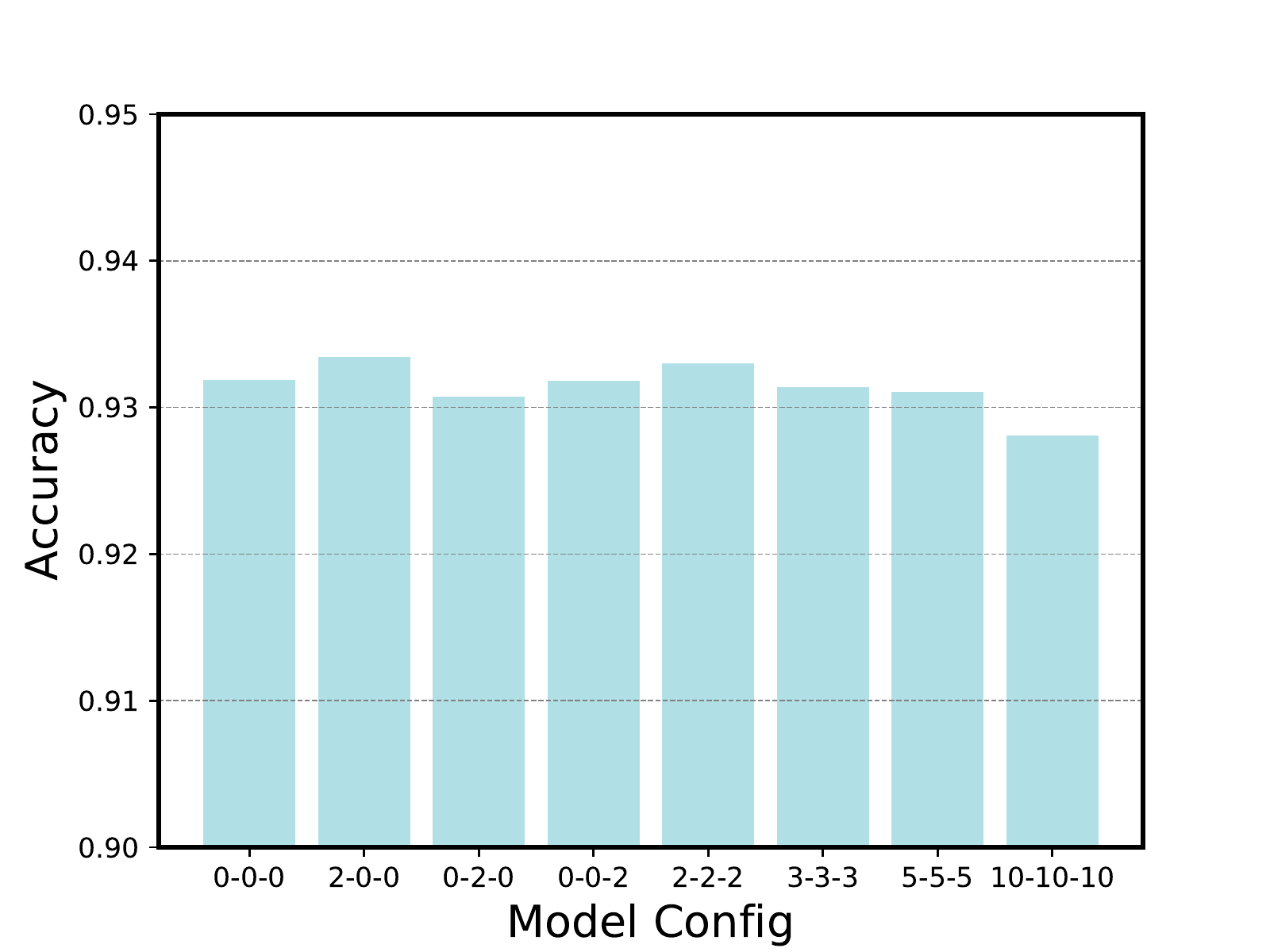}
    \caption{{\small Accuracy for different numbers and positions of prompt tokens on WN11.}}
    \label{fig:ablation}
\end{figure}

We evaluate different numbers and positions of prompt tokens on WN11. We use a sequence $X_1$-$X_2$-$X_3$ to denote the numbers of tokens added in different positions in order. For example, ``2-0-0'' means we add 2 prompt tokens before the head entity and no prompt tokens after the relation and after the tail entity. The results are shown in Figure \ref{fig:ablation}. We observe that ``2-0-0'' performs 
better than ``0-0-0'', and the difference between token numbers and positions is marginal, meaning that what matters is whether to add prompt tokens or not, and numbers and positions are not very important.

\section{Effectiveness of Calibration}
In this section, we show the effectiveness of \calibration. We show the two layers of attention weights of the original BERT and our calibrated \method\ in Figure \ref{fig:cmp_bert}. The left two are attention weights in the middle layer and the right two are in the last layer. For the original BERT, the attention weight in the middle layer between ``type'' and ``clothes'' is small, but it is larger for the \method. And in the last layer, the attention weights of the original BERT between ``\textspt{[CLS]}'' and ``clothes'' and ``type'' are smaller than those of \method. These indicate that our \calibration\ helps to calibrate pretrained LMs for KG completion.



\end{document}